\documentclass{article} 
\usepackage{iclr2027_conference,times}   

\usepackage{amsmath,amsfonts,bm}

\def\eqref#1{equation~\ref{#1}}

\def\1{\bm{1}}

\DeclareMathAlphabet{\mathsfit}{\encodingdefault}{\sfdefault}{m}{sl}
\SetMathAlphabet{\mathsfit}{bold}{\encodingdefault}{\sfdefault}{bx}{n}

\usepackage{graphicx}
\usepackage{float}
\usepackage{booktabs}
\usepackage{multirow}
\usepackage{makecell}
\usepackage{amsmath}
\usepackage{amssymb}
\usepackage{pifont}

\usepackage{url}
\usepackage{hyperref}

\author{Samuel Tetteh \& Cody Fleming \\
Iowa State University\\
Ames, Iowa, USA\\
\texttt{\{samtett, flemingc\}@iastate.edu}}

\title{Safe on Average, Unsafe in the Tail:\\
When Is the Episodic-Cost Tail Controllable?}

\iclrfinalcopy  

\begin{document}
\maketitle

\begin{abstract}
Safe reinforcement learning seeks policies that maximize return while satisfying
constraints on cumulative cost. Most methods impose these constraints on expected
episodic cost. Consequently, standard evaluations report mean episodic cost
without characterizing how cost is distributed across episodes. A policy that
satisfies the mean-cost criterion may therefore remain unsafe in its worst
episodes. Mean-cost reporting neither identifies this tail violation nor shows
whether it can be brought within budget while preserving return. In this work, we
measure the episodic-cost tail using $\mathrm{CVaR}_{0.1}$, the average cost of
the worst $10\%$ of episodes. We classify a policy as tail-safe when
$\mathrm{CVaR}_{0.1}$ is within the safety budget. This allows us first to
identify policies that are safe on average but unsafe in the tail and then to
study whether their tail violations can be controlled while preserving return.
To identify tail-unsafe policies, we evaluate five standard algorithms on three
\emph{Safety-Gymnasium} navigation tasks. We then examine four constraint families on dense-hazard navigation and assess
tail control across four navigation and four locomotion tasks.
\end{abstract}

\section{Introduction}
\begin{figure}[h!]
\centering
\includegraphics[width=0.82\textwidth]{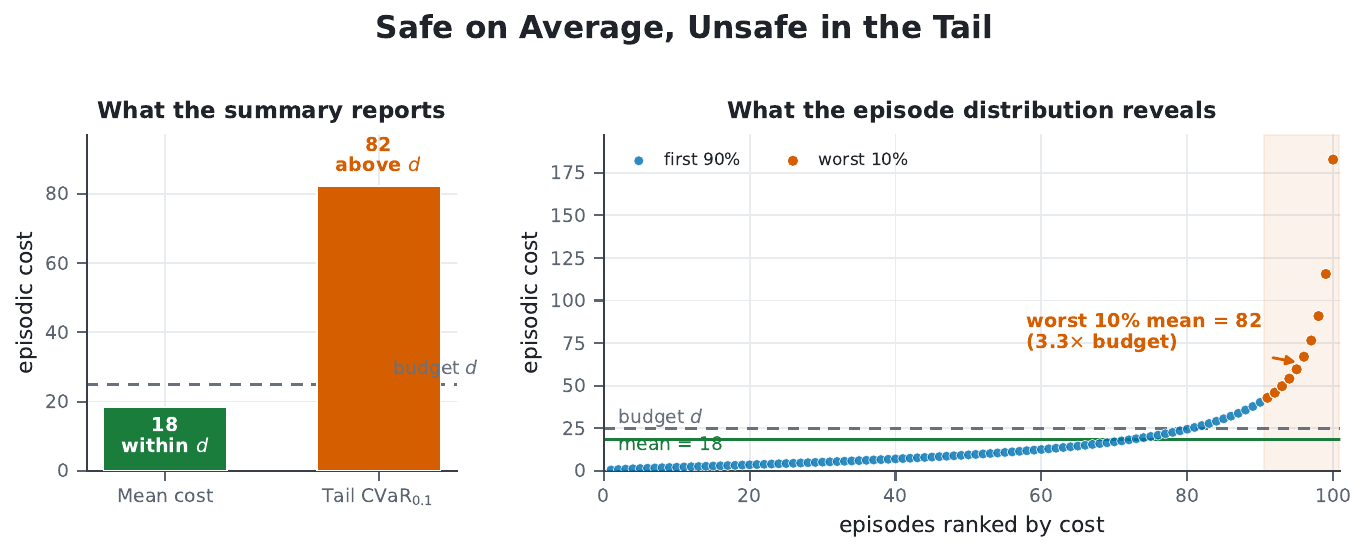}
\caption{\textbf{Mean compliance can hide episodic tail risk.} Both panels describe
the same illustrative policy. Its mean episodic cost is within budget, but the
average cost of its worst $10\%$ of episodes exceeds the budget. Our evaluation
measures both quantities from complete post-training episodes.}
\label{fig:teaser}
\end{figure}

Safe reinforcement learning (RL) seeks policies that maximize return while limiting
undesirable behavior. A nonnegative cost $c(s_t,a_t)$ quantifies undesirable behavior
at time $t$, and the episodic cost $C(\tau)=\sum_t c(s_t,a_t)$ sums this quantity over
the trajectory. Most constrained RL methods require the expected episodic cost to
remain within a budget $d$ \citep{altman1999,achiam2017cpo}. This expectation does
not characterize how cost is distributed across episodes. A policy can satisfy the
mean-cost constraint while a small fraction of episodes incurs costs far above the
budget (Figure~\ref{fig:teaser}).

We measure this high-cost tail using Conditional Value at Risk,
$\mathrm{CVaR}_{\alpha}$, which averages the largest $\alpha$ fraction of episodic
costs \citep{rockafellar2000cvar}. We call a policy tail-safe when
$\mathrm{CVaR}_{0.1}\le d$. Safety-Gymnasium evaluations commonly report aggregate
episodic cost, which does not determine whether this tail criterion is satisfied
\citep{ji2023safetygym}. Although episodic-cost CVaR has been reported on modified
Safety-Gymnasium tasks \citep{orac2025}, no shared evaluation establishes how large
this tail is across standard expected-cost algorithms and tasks.

We study two related questions. First, which policies evaluated using mean cost
remain unsafe in their worst episodes? Second, when can these tail violations be
brought within budget while preserving return? We address the first question through
a common post-training evaluation of five standard safe-RL algorithms on three
navigation tasks. We address the second through $38$ archived training runs from
four constraint families on dense-hazard navigation and an evaluation of tail
control across four navigation and four locomotion tasks.

The initial evaluation reveals substantial tail violations. Across the three
navigation tasks, the average cost of the worst $10\%$ of episodes reaches
$4.4$--$18.2$ times the budget. Strengthening the tested constraints does not resolve
this problem on dense-hazard navigation. Return collapses before the tail reaches the
budget. Locomotion exhibits a different pattern. On the two tasks with the most
severe baseline tails, OQ-SAC, its component ablations, and Lagrangians trained with
lower internal budgets all reach safe operating points without collapsing return.
Tail control is therefore not unique to one method.

The episode-level results suggest an explanation for this difference. Tail control
is observed when a policy can reliably produce episodes that are both safe and
high-return. In the evaluated navigation tasks, safe episodes either have little
return or occur too rarely to control the worst-decile cost. In locomotion, safe
episodes remain frequent while retaining high return. A single-seed hazard-count
analysis in \emph{PointGoal1} examines how the observed return--tail relationship
varies with hazard count.

\paragraph{Contributions.}
Our work makes three contributions.
(1) We apply a shared post-training protocol that measures both mean episodic cost
and $\mathrm{CVaR}_{0.1}$, revealing safety violations that mean-cost reporting does
not identify.
(2) We test four constraint families and document contrasting outcomes for tail
control at held return across the evaluated navigation and locomotion tasks.
(3) We identify the frequency and return of safe episodes as empirical indicators
of tail controllability and examine this interpretation through a within-navigation
hazard-count analysis.

\section{Related Work}
Safe RL is commonly formulated as a constrained Markov decision process
\citep{altman1999,garcia2015survey}. CPO, PPO-Lagrangian, TRPO-Lagrangian,
SAC-Lagrangian, and FOCOPS constrain expected cumulative cost
\citep{achiam2017cpo,ray2019benchmarking,stooke2020responsive,zhang2020focops}.
\emph{Safety-Gymnasium} likewise benchmarks algorithms using aggregate episodic
reward and cost \citep{ji2023safetygym}. These objectives and summaries describe
average constraint satisfaction but do not determine how cost is distributed
across episodes. Our work measures the post-training episode-cost distribution and
asks whether its upper tail can be controlled without reducing return.

Risk-aware safe RL directly optimizes properties of the cost distribution. CVaR
has been used as a risk criterion in RL \citep{tamar2015cvar,chow2017risk}, and
Safe Distributional RL supports CVaR and other constraints on cumulative random
returns \citep{zhang2021sdrl}. WCSAC estimates the distribution of accumulated cost
using a Gaussian model or quantile regression and constrains its CVaR with a
Lagrangian \citep{yang2021wcsac,yang2023distributional}. Trust-region methods form
policy-update surrogates for trajectory-cost CVaR \citep{kim2022offpolicytrc}, while
optimized-certainty-equivalent formulations extend risk-aware objectives beyond
CVaR \citep{lee2025oce}. These works develop methods for optimizing risk-sensitive
constraints. We study when such optimization yields a safe episodic tail without
eliminating task return.

The risk quantities used during training and evaluation are related but not
interchangeable. A distributional critic estimates discounted cost-to-go
conditional on a state and action, while a multiplier may use recent realized
episode costs. Our evaluation instead computes CVaR from the undiscounted costs of
complete post-training episodes. SL-SAC combines an implicit quantile cost critic
with an empirical episodic-CVaR multiplier, although its \emph{Safety-Gymnasium}
tables report mean episodic cost \citep{keswani2026slsac}. ORAC reports
post-training $\mathrm{CVaR}_{0.5}$ and $\mathrm{CVaR}_{0.25}$ on modified
\emph{PointGoal1} and \emph{PointButton1} tasks \citep{orac2025}. We use
$\mathrm{CVaR}_{0.1}$ under one evaluation protocol to characterize the episodic
tail across standard algorithms and tasks.

Other approaches impose safety structure within a trajectory. Saut\'e RL augments
the state with the remaining safety budget to target almost-sure cumulative safety
\citep{sootla2022saute}. Reachability-based methods such as RESPO instead estimate
statewise feasibility \citep{ganai2023respo}. These approaches motivate the
budget-as-state and feasibility-gating families in our constraint study.

Risk-sensitive penalties can also produce conservative exploration. Optimistic
Actor-Critic directs exploration using value uncertainty \citep{ciosek2019oac},
and ORAC extends this idea to risk-averse constrained RL using confidence bounds on
reward and distributional cost critics \citep{orac2025}. OQ-SAC combines this
exploratory action shift with a quantile-CVaR cost critic. Our component ablations
test whether either mechanism is necessary for tail control on \emph{HalfCheetah}
and \emph{Hopper}.

\section{Preliminaries}
\subsection{Constrained reinforcement learning}
We model each task as a finite-horizon constrained Markov decision process (CMDP)
$\mathcal{M}=(\mathcal{S},\mathcal{A},P,r,c,\rho_0,H,d)$ \citep{altman1999}.
The state and action spaces are $\mathcal{S}$ and $\mathcal{A}$. The transition
kernel $P(s'\mid s,a)$ specifies the distribution of the next state $s'$ given
the current state $s$ and action $a$. The functions
$r:\mathcal{S}\times\mathcal{A}\times\mathcal{S}\to\mathbb{R}$ and
$c:\mathcal{S}\times\mathcal{A}\times\mathcal{S}\to\mathbb{R}_{\ge0}$ assign
reward and nonnegative safety cost to each transition. The initial-state
distribution is $\rho_0$, the horizon is $H$, and the episodic cost budget is
$d$.

A policy $\pi(a\mid s)$ generates an episode
$\tau=(s_0,a_0,\ldots,s_T)$, where $s_0\sim\rho_0$,
$a_t\sim\pi(\cdot\mid s_t)$, and $s_{t+1}\sim P(\cdot\mid s_t,a_t)$.
An episode ends when the environment terminates or reaches its time limit, so
$T\le H$. The transition at time $t$ yields reward
$r_t=r(s_t,a_t,s_{t+1})$ and cost $c_t=c(s_t,a_t,s_{t+1})$. The undiscounted
episode totals are
\begin{equation}
R(\tau)=\sum_{t=0}^{T-1} r_t,
\qquad
C(\tau)=\sum_{t=0}^{T-1} c_t.
\label{eq:episode-totals}
\end{equation}
Both task families follow this formulation \citep{ji2023safetygym}. In navigation,
$c_t$ is the sum of the task's enabled binary violation channels at time $t$. In
velocity-constrained locomotion, $c_t$ is one when velocity exceeds the task's
threshold and zero otherwise. The threshold applies to forward velocity in
\emph{HalfCheetah}, \emph{Walker2d}, and \emph{Hopper}, and to planar speed in
\emph{Ant}. All evaluated tasks use $H{=}1{,}000$ and $d{=}25$. The budget applies
to total episode cost, not to an individual transition. The implementations use
discounted targets to train their value functions, while evaluation uses the
undiscounted totals in Eq.~\eqref{eq:episode-totals}.

The initial-state distribution, policy, and transition dynamics induce a
distribution over episodes. Consequently, $R(\tau)$ and $C(\tau)$ may vary across
repeated evaluations. We define the expected return and expected cost as
$J_R(\pi)=\mathbb{E}_{\tau\sim\pi}[R(\tau)]$ and
$J_C(\pi)=\mathbb{E}_{\tau\sim\pi}[C(\tau)]$. The standard CMDP objective is
\begin{equation}
\max_\pi\; J_R(\pi)\qquad\text{subject to}\qquad J_C(\pi)\le d .
\label{eq:cmdp}
\end{equation}
With $d{=}25$, this constraint limits expected episode cost to $25$. It does not
require every episode to remain within budget. For example, a policy that incurs
zero cost in $90$ of $100$ episodes and cost $100$ in the other $10$ has mean cost
$10$. It satisfies the mean-cost constraint even though one episode in ten incurs
four times the budget. Mean cost alone therefore does not characterize the upper
tail of episode cost.

Several baselines in our study optimize this constraint through a Lagrangian
relaxation \citep{ray2019benchmarking,stooke2020responsive}
\begin{equation}
\max_\pi\min_{\lambda\ge0}\;
\mathcal{L}(\pi,\lambda)=J_R(\pi)-\lambda\bigl(J_C(\pi)-d\bigr).
\label{eq:lagrangian}
\end{equation}
The multiplier $\lambda$ penalizes cost in the policy objective. Increasing
$\lambda$ places more weight on reducing cost when the estimated mean exceeds
$d$. This update depends on the scalar violation $J_C(\pi)-d$. Two policies with
the same mean cost therefore produce the same violation even when their episode-cost
distributions differ. Equation~\eqref{eq:lagrangian} places no explicit bound on
the episodic tail. With nonlinear function approximation, the Lagrangian is also an
optimization mechanism and does not guarantee satisfaction of the mean constraint.

\subsection{Episodic tail risk}
For a random episode cost $C$, we parameterize conditional value at risk (CVaR)
by the upper-tail mass $\alpha\in(0,1)$. Thus, $\alpha{=}0.1$ denotes the worst
$10\%$ of episodes. If $F_C^{-1}$ is the generalized quantile function of $C$, then
\begin{equation}
\mathrm{CVaR}_{\alpha}(C)
=\frac{1}{\alpha}\int_{1-\alpha}^{1}F_C^{-1}(u)\,\mathrm{d}u .
\label{eq:cvar}
\end{equation}
This definition handles discrete costs and quantile ties \citep{rockafellar2000cvar}.
Each reported policy is evaluated for $100$ episodes. If
$C_{(1)}\le\cdots\le C_{(100)}$ are the sorted episode costs, the estimator used by
our evaluation code is
\begin{equation}
\widehat{\mathrm{CVaR}}_{0.1}
=\frac{1}{10}\sum_{i=91}^{100} C_{(i)}.
\label{eq:empirical-cvar}
\end{equation}
It is the average cost of the ten most costly episodes. In the example above, it
is $100$, although the mean cost is $10$.

We apply the same budget to the tail and call a policy \emph{tail-safe} when
$\widehat{\mathrm{CVaR}}_{0.1}\le d$. This criterion is stricter than the mean-cost
constraint in Eq.~\eqref{eq:cmdp}, which does not imply tail safety. Tail safety also
does not require every episode to remain within budget. It requires the average of
the worst ten episodes to do so.

Each evaluated policy defines an operating point through its mean return
$\widehat J_R$ and tail cost $\widehat{\mathrm{CVaR}}_{0.1}$, computed from the same
episodes. We assess tail safety together with return because a policy can reduce
cost by ceasing to perform the task. We use \emph{held return} for comparisons
between policies with similar mean return on the same task. The tail is
\emph{controllable at held return} within a tested policy family when its observed
operating points include a tail-safe policy whose return remains comparable
to a higher-tail policy. Since reward scales differ across tasks, we report these
paired measurements without imposing a universal return tolerance.

\section{Episodic-Cost Tails of Standard Safe RL in Navigation}
\label{sec:characterization}
We first evaluate whether standard expected-cost methods control the upper tail of
episodic cost on three navigation tasks. For each available training run, we
evaluate the final policy for $100$ episodes using the deterministic mean action.
These rollouts are independent of training, and each reset samples a new task
layout. We compute mean return, mean cost, and
$\mathrm{CVaR}_{0.1}$ from the raw episode totals for each run, then average the
run-level statistics. Table~\ref{tab:baseline} reports the number of runs used for
each result.

\emph{PointGoal1} provides a common comparison across all five algorithms.
SAC-Lagrangian, TRPO-Lagrangian, and FOCOPS attain similar returns of
$25.4$--$25.7$. Their mean costs of $47$--$52$ already exceed the budget
$d{=}25$, while their tail costs reach $110$--$134$, or $4.4$--$5.4\times d$.
These high-return policies therefore satisfy neither the mean-cost constraint nor
the tail-safety criterion. CPO has an aggregate mean cost of $19$, below the
budget, but its return falls to $2.9$ and its tail remains unsafe at $139$, or
$5.6\times d$. PPO-Lagrangian occupies an intermediate-return point with a still
larger tail of $229$.

Tail violations are larger on \emph{PointGoal2} and \emph{PointButton1}. The
reported policies have tail costs of $246$--$455$, or $9.8$--$18.2\times d$, with
returns no higher than $11.1$. None of the operating points in
Table~\ref{tab:baseline} is tail-safe. Together, these results expose two distinct
failures. High-return policies violate both the mean and tail budgets, while the
lower-mean-cost operating point retains a large tail and low return.
These results motivate a direct test of whether stronger and tail-aware constraints
can reach the tail-safe region at held return.

\begin{table}[t]
\centering
\caption{Episodic-cost tail under the shared post-training evaluation
($d{=}25$, $100$ episodes per run). Statistics are computed for each run and then
averaged. $N$ is the number of available training runs. No reported policy is
tail-safe.}
\label{tab:baseline}
\small
\begin{tabular}{llrrrrr}
\toprule
Algo & Task & $N$ & Return & Mean cost & $\mathrm{CVaR}_{0.1}$ & $\times d$ \\
\midrule
SAC-Lag  & PointGoal1 & 3 & 25.4 & 48 & 116 & 4.6 \\
TRPO-Lag & PointGoal1 & 3 & 25.7 & 47 & 110 & 4.4 \\
FOCOPS   & PointGoal1 & 3 & 25.4 & 52 & 134 & 5.4 \\
PPO-Lag  & PointGoal1 & 4 & 10.2 & 55 & 229 & 9.2 \\
CPO      & PointGoal1 & 3 &  2.9 & 19 & 139 & 5.6 \\
\midrule
TRPO-Lag & PointGoal2 & 3 & 7.9 & 87 & 350 & 14.0 \\
SAC-Lag  & PointGoal2 & 1 & $-0.0$ & 51 & 455 & 18.2 \\
FOCOPS   & PointButton1 & 3 & 11.1 & 86 & 246 & 9.8 \\
TRPO-Lag & PointButton1 & 2 & 5.2 & 79 & 353 & 14.1 \\
\bottomrule
\end{tabular}
\end{table}

\section{Constraint Tightening Fails at Held Return in Dense-Hazard Navigation}
\label{sec:negative}
\begin{figure}[t]
\centering
\includegraphics[width=0.74\columnwidth]{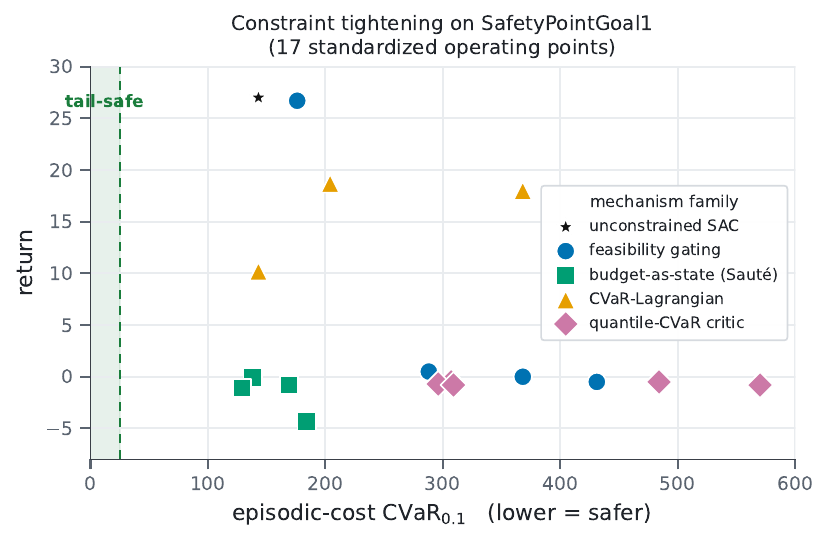}
\caption{The negative result on \emph{PointGoal1} ($d{=}25$). The figure shows
$17$ operating points, including an unconstrained reference. Each point plots return
against episodic-cost $\mathrm{CVaR}_{0.1}$. Constraint tightening collapses return
without bringing the tail within the shaded safe region.}
\label{fig:negative}
\end{figure}

To determine whether this failure is specific to standard expected-cost objectives,
we test four stronger or tail-aware constraint families. Figure~\ref{fig:negative}
reports the $17$ \emph{PointGoal1} operating points with standardized post-training
measurements. The complete archive contains $38$ training runs across
\emph{PointGoal1} and \emph{PointButton1}.
\begin{itemize}
\setlength{\itemsep}{0pt}
\setlength{\parsep}{0pt}
\item \textbf{Feasibility gating.}
We learn a max-over-time reach-avoid critic $V_h(s,a)$. Its target is the larger
of the current violation indicator and the discounted next-state value. We use
$V_h$ with scalar and state-dependent multipliers and with a RESPO-style gate
\citep{ganai2023respo}. The gate maximizes return in states predicted to be
feasible and minimizes $V_h$ elsewhere. Thirteen configurations vary the gating
rule, multiplier structure, and feasibility threshold.
\item \textbf{Budget as state.}
Following Saut\'e RL \citep{sootla2022saute}, we append a normalized budget balance
$z_t$ to the state. It starts at $z_0=1$ and follows
$z_{t+1}=z_t-c_t/d$, so $z_t$ is the fraction of the episode budget not yet
spent. It becomes negative when cumulative cost exceeds $d$. On steps with
$z_t<0$, the task reward is replaced by a negative penalty. Four budget and
penalty settings test whether explicit episode-level budget memory prevents large
costs.
\item \textbf{Tail-driven Lagrangian.}
We retain the scalar cost critic and actor objective of SAC-Lagrangian but update
the multiplier from the costs of the ten most recently completed training
episodes. We use either empirical $\mathrm{CVaR}_{0.1}$ or the surrogate
$\bar C+k\sigma_C$ \citep{yang2021wcsac,keswani2026slsac}. A replay variant forms
roughly one quarter of each minibatch by sampling from the $10\%$ of stored
transitions with the highest observed one-step costs. This family tests whether a
tail-aware multiplier and replay distribution suffice without modeling the
cost-return distribution at each state-action pair.
\item \textbf{Per-state quantile-CVaR critic.}
We replace the scalar cost critic with a 32-quantile model of the cost return
$Z^C(s,a)$ trained by quantile regression \citep{dabney2018qrdqn}. The actor
penalty averages the upper $\alpha$ fraction of the predicted quantiles, while
the multiplier tracks empirical episodic $\mathrm{CVaR}_{0.1}$. During critic
training, one quarter of each minibatch is drawn from the $10\%$ of stored
transitions with the highest observed one-step costs. This family tests whether
a state-action-specific tail estimate can penalize high-risk decisions while
preserving reward elsewhere.
\end{itemize}
All four families exhibit the same failure. As each method penalizes cost more strongly,
return approaches zero, while $\mathrm{CVaR}_{0.1}$ remains above approximately
$130$ and often rises. Figure~\ref{fig:negative} shows this pattern across $17$
operating points, including an unconstrained reference. None reaches the tail-safe
region. Within multiplier-based families, only settings with
$\lambda\approx0$ retain high return, and their mean costs far exceed $d$. This
result does not isolate its cause. One possible explanation is that hazards occupy
task-relevant states, limiting the selectivity of both global and state-specific
penalties. Appendix~\ref{app:negative-configs} summarizes the full study, and
Figures~\ref{fig:constraint-diagnostics} and~\ref{fig:constraint-replication} report
all $38$ preserved training traces for \emph{PointGoal1} and \emph{PointButton1}.
The quantile-CVaR and tail-driven Lagrangian replications on \emph{PointButton1}
reproduce this failure. Return remains near zero at every tightening level, while
$\mathrm{CVaR}_{0.1}$ remains between $200$ and $420$, equivalent to $8$--$17$
times the budget.

\section{Optimistic Quantile-CVaR SAC}
\label{sec:method}
To study tail control when high-return and low-cost behavior may be separated, we use
an SAC controller adapted from ORAC \citep{haarnoja2018sac,orac2025}. We call it
\emph{OQ-SAC}, short for optimistic quantile-CVaR SAC. It combines optimistic data
collection with a quantile-CVaR penalty for target-policy learning. OQ-SAC is a
test case for reaching a safe, high-return operating point. Table~\ref{tab:ablation}
shows that the optimistic mean-cost variant and the
quantile-CVaR variant without the action shift both reach safe operating points on
the evaluated locomotion tasks.

\paragraph{Relationship to ORAC.} ORAC combines reward critics, an ensemble of
quantile cost critics, a CVaR actor penalty, and a separate exploratory policy based
on confidence bounds \citep{orac2025}. It shifts the target-policy mean using a
covariance-scaled gradient whose size is controlled by a KL-divergence bound. Our
implementation retains the confidence-bound objective but applies a normalized
gradient shift to a sampled action. It also samples costly transitions more often
during critic training and holds the multiplier fixed for each task. These changes
define the controller evaluated here. The quantile critics and optimistic
exploration are established components. We use their combination to study when tail
control is possible and test each component separately.

\paragraph{Ensemble critics.} Each of the $E$ reward critics predicts
$q_e(s,a)=Q^R_e(s,a)$, and each cost critic predicts $N$ quantiles of the
discounted cost return using quantile regression \citep{dabney2018qrdqn}. We define
$c_e(s,a)=\mathrm{CVaR}_\alpha(Z^C_e(s,a))$, where the empirical CVaR is the
average of the largest $\lceil\alpha N\rceil$ predicted quantiles. The confidence
bounds used for data collection are
\begin{align}
\widehat{Q}^R(s,a) &= \bar q(s,a) + \beta_R s_q(s,a),\\
\widehat{Q}^C(s,a) &= \bar c(s,a) - \beta_C s_c(s,a),
\end{align}
where $\bar q$ and $\bar c$ are ensemble means, while $s_q$ and $s_c$ are the
corresponding standard deviations across critics.

\paragraph{Exploratory action.} After the initial random data collection, the
target policy samples $a_T\sim\pi(\cdot\mid s)$. We compute
\[
g_t=\nabla_a\!\left[\widehat Q^R(s,a)-\lambda\widehat Q^C(s,a)\right]_{a=a_T}
\]
and execute
\[
a_E=\operatorname{clip}\!\left(
a_T+\delta\,\mathbf a_{\max}\odot\frac{g_t}{\lVert g_t\rVert_2+\varepsilon},
-\mathbf a_{\max},\mathbf a_{\max}\right).
\]
Here $\mathbf a_{\max}$ contains the action limits and $\varepsilon$ is a small
numerical constant. Normalizing the gradient makes $\delta$ control the step size
in normalized action coordinates. Clipping enforces the environment's action
bounds. This shift is used only for data collection. The target policy is optimized
with the objective below.
Figure~\ref{fig:method} gives a geometric interpretation of this action shift in
Appendix~\ref{app:schematic}.

\paragraph{Target-policy loss and scaling.} The target policy minimizes
\[
\mathcal L_\pi=
\mathbb E_{s\sim\mathcal D,\,a\sim\pi}\!\left[
\alpha_{\mathrm{ent}}\log\pi(a\mid s)-\bar q(s,a)+\lambda\bar c(s,a)
\right]
\]
using the ensemble means defined above. The confidence bonuses affect data
collection but do not enter this loss. The cost term gives a larger penalty to
actions with larger predicted tail cost at the current state.

We hold $\lambda$ fixed for each task. The reported settings use
$\lambda\!\approx\!0.06$--$0.15$ on navigation and
$\lambda\!\approx\!20$--$30$ on velocity because reward values are of order tens
in navigation and thousands in locomotion. A reactive update was unstable in a
single-seed Ant comparison, as reported in Appendix~\ref{app:hparams}. During critic
training, one quarter of each minibatch is drawn from the $10\%$ of stored
transitions with the largest observed one-step costs. This sampling rule increases
the frequency of costly transitions during quantile-critic updates.

\section{Experiments}
\label{sec:experiments}
We evaluate OQ-SAC on four velocity-control tasks and four dense-hazard
navigation tasks from Safety-Gymnasium \citep{ji2023safetygym}. The locomotion
tasks are \emph{HalfCheetah}, \emph{Ant}, \emph{Walker2d}, and \emph{Hopper}.
The navigation tasks are \emph{PointGoal2}, \emph{PointButton1},
\emph{CarButton1}, and \emph{PointPush1}.

The primary baselines are SAC-Lagrangian and PPO-Lagrangian from \emph{OmniSafe}.
Training uses $10^6$ environment interactions, except for OQ-SAC on
\emph{HalfCheetah}, which uses $5\times10^5$. Each final policy is evaluated for
$100$ deterministic episodes with budget $d{=}25$. Results use three seeds except
OQ-SAC on \emph{Walker2d}, which uses five. A method is classified as safe only
if $\mathrm{CVaR}_{0.1}\!\le\!d$ on every seed.

\subsection{Locomotion results}
OQ-SAC is safe on every seed for \emph{HalfCheetah}, \emph{Ant}, and
\emph{Hopper}. It is safe on four of five \emph{Walker2d} seeds, with the
remaining seed at $28.3$. Among the methods in Table~\ref{tab:main}, OQ-SAC is
the only one safe on every seed for both \emph{HalfCheetah} and \emph{Hopper}.
SAC-Lagrangian is safe on \emph{Ant} and \emph{Walker2d}. It has higher return
on \emph{Ant}, while OQ-SAC has higher mean return on \emph{Walker2d} but misses
the seedwise safety criterion. PPO-Lagrangian is unsafe on at least one seed in
all four tasks.

Lower internal cost targets also produce safe Lagrangian policies in a
single-seed comparison, and WCSAC is safe on \emph{HalfCheetah} and \emph{Ant}
but not on \emph{Walker2d} or \emph{Hopper}. Appendix~\ref{app:sweep} and
Appendix~\ref{app:wcsac} report these comparisons. They show that safe locomotion
tails are not unique to OQ-SAC.

\begin{table*}[t]
\centering
\caption{Performance after $100$ deterministic evaluation episodes with
$d{=}25$. Training uses $10^6$ interactions except OQ-SAC on
\emph{HalfCheetah}, which uses $5\times10^5$. Return and CVaR are means and
standard deviations across seeds, while cost is the mean episodic cost. Results
use three seeds except OQ-SAC on \emph{Walker2d}, which uses five. Bold CVaR
entries are safe on every seed. \underline{Underline} marks the highest-return
safe result. $^{\dagger}$OQ-SAC is safe on four of five \emph{Walker2d} seeds,
with the remaining seed at $\mathrm{CVaR}_{0.1}{=}28.3$.}
\label{tab:main}
\setlength{\tabcolsep}{5pt}
\resizebox{\textwidth}{!}{%
\begin{tabular}{l ccc ccc ccc}
\toprule
& \multicolumn{3}{c}{\textbf{OQ-SAC (ours)}} & \multicolumn{3}{c}{SAC-Lag} & \multicolumn{3}{c}{PPO-Lag} \\
\cmidrule(lr){2-4}\cmidrule(lr){5-7}\cmidrule(lr){8-10}
Task & $R\!\uparrow$ & cost & $\mathrm{CVaR}\!\downarrow$ & $R\!\uparrow$ & cost & $\mathrm{CVaR}\!\downarrow$ & $R\!\uparrow$ & cost & $\mathrm{CVaR}\!\downarrow$ \\
\midrule
\multicolumn{10}{l}{\textit{Locomotion}}\\
HalfCheetah  & \underline{$2868_{\pm46}$} & 0 & $\boldsymbol{0}$ & $3255_{\pm2757}$ & 308 & $597_{\pm425}$ & $3146_{\pm288}$ & 548 & $581_{\pm185}$ \\
Hopper       & \underline{$877_{\pm556}$} & 0.1 & $\boldsymbol{1.0_{\pm1.3}}$ & $1164_{\pm222}$ & 70 & $97_{\pm137}$ & $517_{\pm597}$ & 7.5 & $18_{\pm18}$ \\
Ant          & $2656_{\pm329}$ & 0.9 & $\boldsymbol{0.9_{\pm0.6}}$ & \underline{$2909_{\pm40}$} & 2.9 & $\boldsymbol{6.1_{\pm2.0}}$ & $171_{\pm92}$ & 4.1 & $15_{\pm8}$ \\
Walker2d     & $1617_{\pm949}$ & 2.4 & $10.7_{\pm10.1}^{\dagger}$ & \underline{$1273_{\pm1175}$} & 1.6 & $\boldsymbol{6.8_{\pm9.6}}$ & $551_{\pm129}$ & 52 & $54_{\pm41}$ \\
\midrule
\multicolumn{10}{l}{\textit{Navigation}}\\
PointGoal2   & $4.6_{\pm4.3}$ & 58 & $358_{\pm51}$ & $-0.2_{\pm0.1}$ & 55 & $475_{\pm111}$ & $5.5_{\pm0.8}$ & 90 & $297_{\pm47}$ \\
PointButton1 & $1.2_{\pm1.3}$ & 99 & $397_{\pm111}$ & $-2.6_{\pm3.1}$ & 45 & $245_{\pm65}$ & $7.0_{\pm1.1}$ & 71 & $242_{\pm51}$ \\
CarButton1   & $-0.7_{\pm0.6}$ & 88 & $468_{\pm24}$ & $-0.3_{\pm0.1}$ & 49 & $331_{\pm70}$ & $4.0_{\pm0.7}$ & 143 & $419_{\pm31}$ \\
PointPush1   & $-0.1_{\pm0.2}$ & 20 & $204_{\pm139}$ & $-0.0_{\pm0.2}$ & 8 & $76_{\pm18}$ & $0.6_{\pm0.1}$ & 49 & $396_{\pm89}$ \\
\bottomrule
\end{tabular}%
}
\end{table*}

\subsection{Navigation results}
No method in Table~\ref{tab:main} satisfies the tail constraint on a navigation
task. Mean return never exceeds $8$, while $\mathrm{CVaR}_{0.1}$ ranges from
$76$ to $475$. On \emph{PointPush1}, OQ-SAC has mean cost $20$ but
$\mathrm{CVaR}_{0.1}{=}204$, illustrating why mean cost does not establish tail
safety. None of the evaluated methods controls the tail on these dense-hazard
navigation tasks.

\subsection{Component analysis}
On \emph{HalfCheetah} and \emph{Hopper}, both the mean-cost variant and the
quantile-CVaR variant without the action shift are safe on every seed. The
training-budget-matched \emph{Hopper} ablations also obtain higher mean return
than full OQ-SAC. HalfCheetah returns are not directly comparable because the
full method uses fewer training interactions. Neither component is individually
necessary for tail safety on these two tasks, although we do not remove both
simultaneously. All variants use a task-specific fixed multiplier, so this
ablation does not isolate multiplier selection. Table~\ref{tab:ablation} in
Appendix~\ref{app:ablations} reports the results. Appendix~\ref{app:hparams}
describes the fixed and reactive multiplier comparison.

\section{When Is the Tail Controllable?}\label{sec:when}
\begin{figure}[t]
\centering
\includegraphics[width=0.82\textwidth]{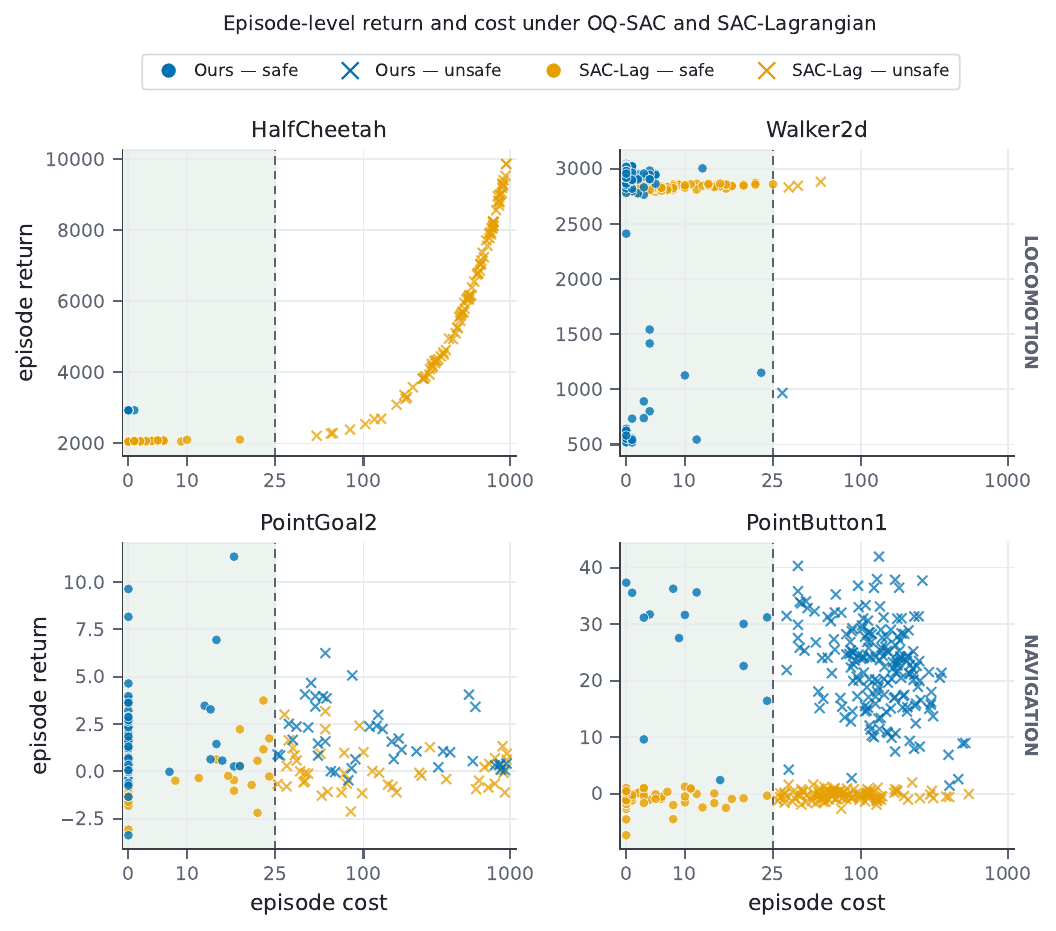}
\caption{Episode return and cost over $200$ evaluations per policy. Circles are
safe episodes with $C(\tau)\!\le\!d$, and crosses are unsafe. Safe, high-return
episodes are frequent for the shown locomotion policies and absent or infrequent
for the shown navigation policies.}
\label{fig:classes}
\end{figure}

The experiments suggest two episode-level signals for tail controllability. The
first is the cost distribution. A representative SAC-Lagrangian policy on
\emph{HalfCheetah} has separated low-cost and high-cost clusters, whereas costs
on \emph{PointButton1} are more diffuse (Figure~\ref{fig:separability}).
The second is whether safe episodes occur frequently while retaining high return.
Figure~\ref{fig:classes} shows that OQ-SAC is safe on $100\%$ of
\emph{HalfCheetah} episodes and $99.5\%$
of \emph{Walker2d} episodes, with median safe returns of $2931$ and $2962$. Safe
episodes on \emph{PointGoal2} have near-zero return. On \emph{PointButton1},
OQ-SAC attains high return in some safe episodes, but they form only $7\%$ of the
evaluation. A safe, high-return outcome must therefore occur often enough to
control the tail.

Across the eight tasks, these signals agree with the observed results. Several
methods find safe, high-return locomotion policies, while none satisfies the tail
constraint on navigation. This is an empirical association, not a necessary or
sufficient condition for tail control.

Table~\ref{tab:taxonomy} summarizes these episode-level patterns.
\begin{table}[t]
\centering
\caption{Episode-level patterns associated with tail control in the evaluated
tasks. These patterns summarize the experiments and are not necessary or
sufficient conditions. The single-seed hazard-count study in Table~\ref{tab:density}
provides a within-navigation sensitivity analysis.}
\label{tab:taxonomy}
\small
\begin{tabular}{lcc}
\toprule
Observed evidence & Locomotion & Navigation \\
\midrule
Representative cost example & Separated clusters (HalfCheetah) & Diffuse costs (PointButton1) \\
Safe, high-return episodes in Figure~\ref{fig:classes} & Frequent & Absent or infrequent \\
\midrule
Safe tail at held return found & Yes & No \\
OQ-SAC safe on every seed & 3 of 4 tasks & 0 of 4 tasks \\
\bottomrule
\end{tabular}
\end{table}

\paragraph{Mean and tail cost.} The worst-decile cost reaches $10.2$ times the
mean in our results (Table~\ref{tab:meanvstail}). A moderate mean can therefore
hide a severe tail. A \emph{HalfCheetah} multiplier sweep reaches the safe band
with return near $2900$, whereas none of the evaluated \emph{PointGoal1}
configurations enters it (Figure~\ref{fig:decouple}).

\paragraph{Within-domain sensitivity.} We vary the \emph{PointGoal1} hazard count
with one seed while holding the OQ-SAC configuration, budget, and
$\lambda{=}0.06$ fixed. At one, four, and eight hazards, return stays near $27$
while $\mathrm{CVaR}_{0.1}$ rises from $40.9$ to $82.3$ and $176.9$. At sixteen
hazards, return falls to $6.8$ and CVaR reaches $397.2$. The two-hazard result is
non-monotonic, and no setting is safe. The observed return--tail operating point
varies with hazard count, but this single-seed comparison cannot establish a
systematic effect.
Appendix~\ref{app:density} reports all values.

\section{Discussion and Limitations}
Across the evaluated tasks, tail control is observed only when policies reliably
produce episodes that are both safe and high-return. OQ-SAC satisfies the seedwise
tail constraint in three of four locomotion tasks and none of the four navigation tasks. This pattern is
not unique to OQ-SAC. Both ablated variants are safe on \emph{HalfCheetah} and
\emph{Hopper}, while stricter expected-cost targets produce safe policies in
single-seed comparisons. The evidence supports the episode-level interpretation in
Section~\ref{sec:when}, but does not establish that a particular algorithm or
component is necessary.

Domain and observed cost structure remain confounded. Every task in which a
method controls the tail is a locomotion task, while every task in which none does
is a navigation task. The one-seed hazard-count study finds no safe setting, so
it neither establishes a causal effect nor identifies a transition from a safe to
an unsafe setting. OQ-SAC also uses task-specific fixed multipliers. Each main
per-seed
$\mathrm{CVaR}_{0.1}$ estimate averages ten tail episodes from $100$
deterministic-policy evaluations, and most results use three training seeds. The
experiments use low-dimensional observations and one cost signal, leaving image
observations and multiple simultaneous constraints untested. These limitations
motivate reporting the episodic-cost tail together with the frequency and return
of safe episodes to show whether a method controls rare failures without
sacrificing return.

\subsubsection*{AI use statement}
We used a generative AI assistant to assist in revising the manuscript. The authors take responsibility for the final content of the manuscript submitted for review.

\bibliographystyle{iclr2027_conference}
\bibliography{tail_method}

\appendix

\section{Compute and Reproducibility}
All experiments run on CPU nodes under SLURM. Each run trains for $10^6$
environment interactions, except OQ-SAC on \emph{HalfCheetah}, which trains for
$5\times10^5$. Each final policy is evaluated for $100$ episodes with the
deterministic mean action. Evaluation episode $i$ is reset with seed $1000+i$, and
return and cost accumulate until termination or truncation.
$\mathrm{CVaR}_{0.1}$ follows Eq.~\eqref{eq:empirical-cvar}.

Results average three training seeds, except OQ-SAC on \emph{Walker2d}, which
uses five. We report the mean and the population standard deviation across seeds
(denominator equal to the number of seeds). The software stack is Python 3.10.20,
\emph{PyTorch} 2.3.1, \emph{Gymnasium} 0.28.1, and \emph{Safety-Gymnasium} 1.0.0
\citep{ji2023safetygym}. Code, launch scripts, and saved configurations will be
released with the paper.

\section{Additional Episodic-Tail Measurements}

\subsection{Relation to Prior Tail-Risk Evaluation}
\label{app:gap}

Prior work has used CVaR and other distributional risk measures to constrain
cumulative cost during training
\citep{zhang2021sdrl,yang2021wcsac,kim2022offpolicytrc}. ORAC also reports
episodic-cost $\mathrm{CVaR}_{0.5}$ and $\mathrm{CVaR}_{0.25}$ for
SAC-Lagrangian, WCSAC, and ORAC on modified \emph{PointGoal1} and
\emph{PointButton1} tasks with a 400-step horizon and budget $d{=}10$
\citep{orac2025}. SL-SAC uses empirical episodic-cost CVaR to update its
multiplier, while reporting mean episodic cost in its main benchmark results
\citep{keswani2026slsac}.

We use $\mathrm{CVaR}_{0.1}$ as a shared post-training measure with a
1000-step horizon and budget $d{=}25$. Applying the same evaluation to five
standard mean-constrained algorithms reveals tail violations on three navigation
tasks. We then use this measure to study tail controllability across four
navigation and four locomotion tasks.

\subsection{Mean Cost Versus Episodic Tail}
\label{app:meanvstail}

Table~\ref{tab:meanvstail} compares mean episode cost with
$\mathrm{CVaR}_{0.1}$ for OQ-SAC and SAC-Lagrangian on all eight tasks. The
tail-to-mean ratio measures how much larger the average cost of the worst
$10\%$ of episodes is than the overall episode mean.
\begin{table}[H]
\centering
\caption{Mean episode cost and $\mathrm{CVaR}_{0.1}$ under the shared
$100$-episode deterministic evaluation with $d{=}25$. The tail-to-mean ratio
compares the average cost of the worst $10\%$ of episodes with the mean across
all episodes. Navigation entries are averaged across three training seeds. A
dash indicates that both cost statistics are zero.}
\label{tab:meanvstail}
\small
\begin{tabular}{l ccc ccc}
\toprule
& \multicolumn{3}{c}{\textbf{OQ-SAC (ours)}} & \multicolumn{3}{c}{SAC-Lagrangian} \\
\cmidrule(lr){2-4}\cmidrule(lr){5-7}
Task & mean & $\mathrm{CVaR}_{0.1}$ & tail/mean & mean & $\mathrm{CVaR}_{0.1}$ & tail/mean \\
\midrule
\multicolumn{7}{l}{\textit{Locomotion}}\\
HalfCheetah  & $0.0$ & $0$   & --          & $308$ & $597$ & $1.9\times$ \\
Hopper       & $0.1$ & $1.0$ & $10.0\times$ & $70$  & $97$  & $1.4\times$ \\
Ant          & $0.9$ & $0.9$ & $1.0\times$  & $2.9$ & $6.1$ & $2.1\times$ \\
Walker2d     & $2.4$ & $10.7$& $4.5\times$  & $1.6$ & $6.8$ & $4.3\times$ \\
\midrule
\multicolumn{7}{l}{\textit{Navigation}}\\
PointGoal2   & $58$ & $358$ & $6.2\times$  & $55$ & $475$ & $8.6\times$ \\
PointButton1 & $99$ & $397$ & $4.0\times$  & $45$ & $245$ & $5.4\times$ \\
CarButton1   & $88$ & $468$ & $5.3\times$  & $49$ & $331$ & $6.8\times$ \\
PointPush1   & $20$ & $204$ & $10.2\times$ & $8$  & $76$  & $9.5\times$ \\
\bottomrule
\end{tabular}
\end{table}

\subsection{Within-Episode Cost Trajectories}

\begin{figure}[ht]
\centering
\includegraphics[width=\textwidth]{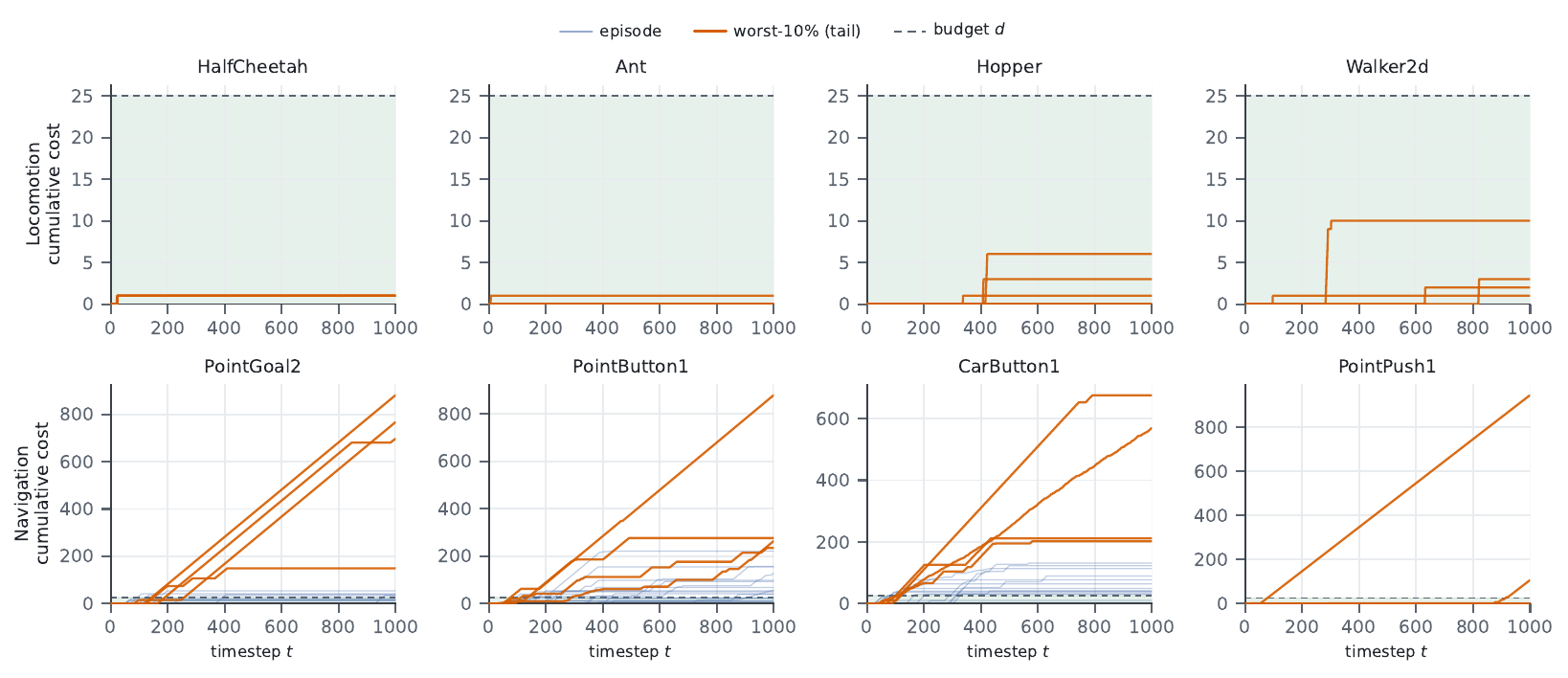}
\caption{Cumulative cost over $40$ deterministic episodes from one OQ-SAC
policy per task. Orange curves identify the four episodes with the largest
final costs. Blue curves show the remaining episodes, and the dashed line marks
the episodic budget $d{=}25$.}
\label{fig:trajectories}
\end{figure}

Figure~\ref{fig:trajectories} shows when cost is incurred within an episode.
For one OQ-SAC policy per task, we run $40$ deterministic episodes using
evaluation seeds $4000$ through $4039$. Each curve shows the cumulative cost
$\sum_{j=0}^{t}c_j$ and remains constant after termination. The four episodes
with the largest final costs form the empirical worst $10\%$ and are
highlighted. Costs remain near zero for the selected locomotion policies. The
highest-cost navigation episodes accumulate violations across substantial
portions of the rollout.

\section{Configuration Coverage and Training Trajectories}
\label{app:negative-configs}

\begin{figure}[H]
\centering
\includegraphics[width=\textwidth]{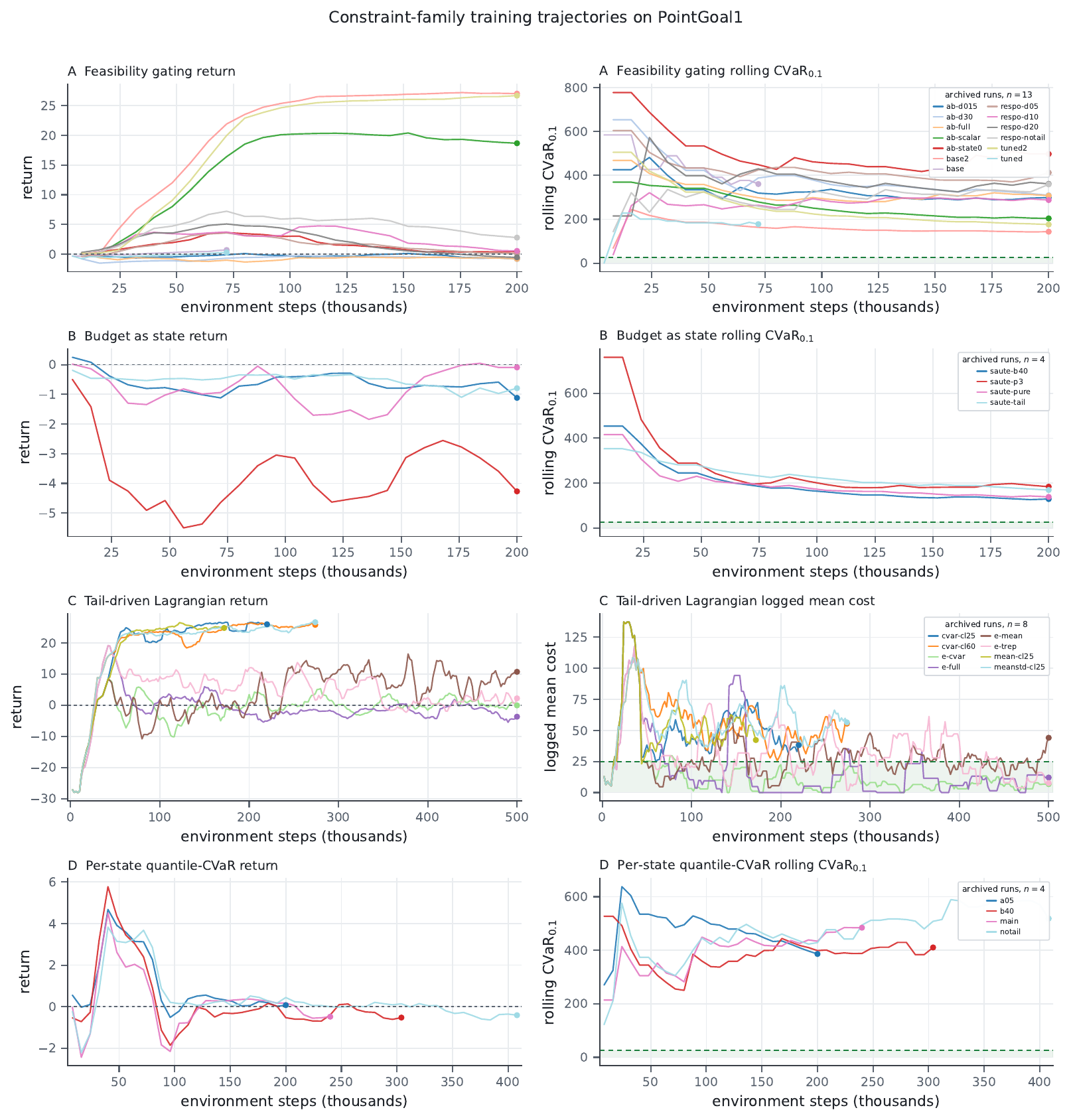}
\caption{Training trajectories for the $29$ archived \emph{PointGoal1} runs.
Panels A, B, and D show logged return and rolling empirical
$\mathrm{CVaR}_{0.1}$. Panel C shows deterministic test return and mean episode
cost because the \emph{OmniSafe} logs do not contain the episode-level costs
required to reconstruct CVaR. Each line represents one configuration, and each
dot marks its final logged value. The shaded green region marks costs at or
below $d{=}25$. These training statistics are separate from the shared
post-training evaluation.}
\label{fig:constraint-diagnostics}
\end{figure}

\begin{figure}[H]
\centering
\includegraphics[width=\textwidth]{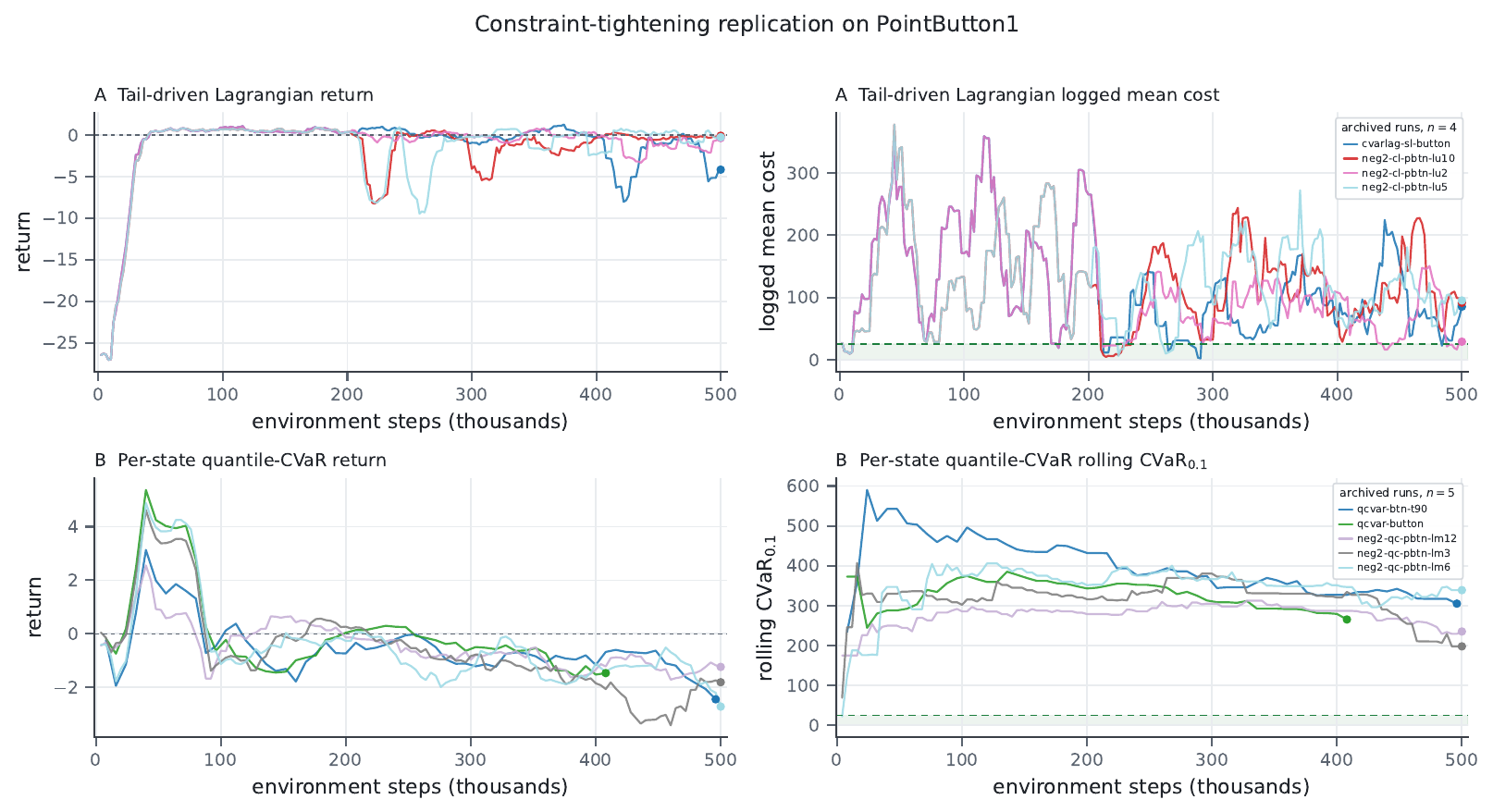}
\caption{Training trajectories for the nine archived \emph{PointButton1} runs.
Panel A shows deterministic test return and mean episode cost for four
tail-driven Lagrangian configurations. Panel B shows logged return and rolling
empirical $\mathrm{CVaR}_{0.1}$ for five per-state quantile-CVaR
configurations. Each dot marks the final logged value. The shaded green region
marks costs at or below $d{=}25$.}
\label{fig:constraint-replication}
\end{figure}

The run archive contains $38$ single-seed training traces from the four
constraint families evaluated in Section~\ref{sec:negative}. Of these, $29$ use
\emph{PointGoal1} and $9$ use \emph{PointButton1}.
Table~\ref{tab:negconfigs} reports the number of configurations retained from
each family. Figures~\ref{fig:constraint-diagnostics}
and~\ref{fig:constraint-replication} show how return and the available cost
statistic evolved during training. These trajectories complement the shared
post-training evaluation used for the conclusions in
Section~\ref{sec:negative}.

\begin{table}[H]
\centering
\small
\begin{tabular}{lccc}
\toprule
Mechanism family & PointGoal1 & PointButton1 & Total \\
\midrule
Feasibility gating & 13 & 0 & 13 \\
Budget as state & 4 & 0 & 4 \\
Tail-driven Lagrangian & 8 & 4 & 12 \\
Per-state quantile-CVaR critic & 4 & 5 & 9 \\
\midrule
Total & 29 & 9 & 38 \\
\bottomrule
\end{tabular}
\caption{Archived training traces from the constraint-mechanism study. Each
trace corresponds to one method configuration and one training seed.}
\label{tab:negconfigs}
\end{table}

\section{Method Details}

\subsection{OQ-SAC Implementation and Hyperparameters}
\label{app:hparams}

The actor and each reward and cost critic use two hidden layers of $256$ ReLU
units. The actor is a squashed Gaussian policy whose log standard deviation is
clipped to $[-20,2]$. Adam updates the actor, critics, and learned entropy
temperature with learning rate $3\times10^{-4}$. The initial entropy coefficient
is $0.2$, and the target entropy is $-|\mathcal{A}|$. We use
$\gamma=\gamma_c=0.99$, a target-network update coefficient of $0.005$, batch
size $256$, and replay capacity $5\times10^5$. Each run begins with $10^4$
random interactions. True environment terminations stop Bellman
bootstrapping, while time-limit truncations do not.

Every reported OQ-SAC policy uses three reward critics and three cost critics.
Each cost critic predicts $16$ quantiles at the midpoints of uniformly spaced
quantile intervals and is trained with the quantile Huber loss. The actor
penalty averages the four largest predicted quantiles, corresponding to
$\alpha_{\mathrm{pen}}=0.25$. One quarter of each critic minibatch is sampled
from the $10\%$ of stored transitions with the largest observed one-step costs.
The confidence-bound parameters are $\beta_R=2$ and $\beta_C=1$, and the
normalized exploratory action shift uses $\delta=0.03$.

We hold the multiplier fixed within each task. We use $\lambda=25$ for
\emph{HalfCheetah}, $20$ for \emph{Ant} and \emph{Walker2d}, and $30$ for
\emph{Hopper}. We use $\lambda=0.06$ for \emph{PointGoal2} and $0.15$ for
\emph{PointButton1}, \emph{CarButton1}, and \emph{PointPush1}. The smaller
navigation multipliers reflect the lower reward and cost scales of those tasks.

\subsection{Fixed and Reactive Multiplier Comparison}

In a single-seed \emph{Ant} run, the reactive multiplier produced a logged
training return of $2715$ at $688{,}000$ interactions. Its final deterministic
policy obtained return $-628$ and $\mathrm{CVaR}_{0.1}=6.1$. Across three seeds,
the fixed-multiplier policies obtained return $2656_{\pm329}$ and
$\mathrm{CVaR}_{0.1}=0.9_{\pm0.6}$. This comparison motivated the fixed
multipliers used in our experiments but does not establish that fixed
multipliers are generally preferable.

\subsection{Conceptual View of the Exploratory Action Shift}
\label{app:schematic}

\begin{figure}[H]
\centering
\includegraphics[width=0.60\textwidth]{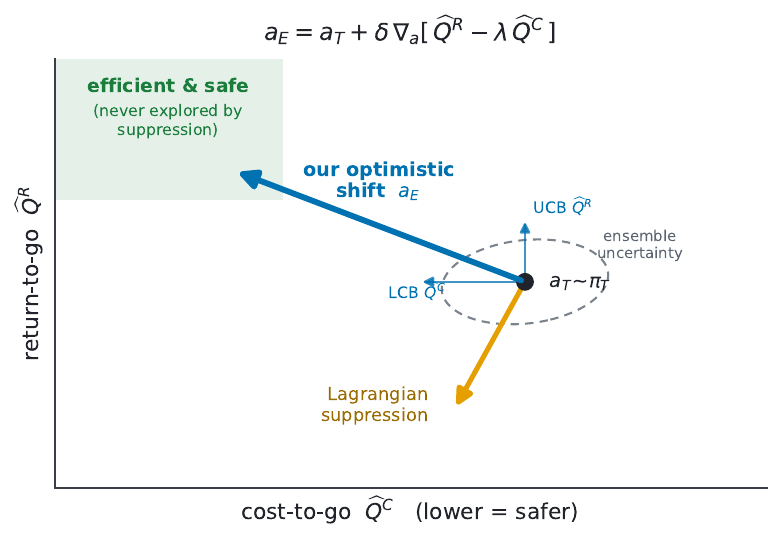}
\caption{Conceptual projection of the exploratory action shift. The blue arrow
represents the change from the target-policy action $a_T$ to the exploratory
action $a_E$. This update follows the action-space gradient of the
upper-confidence reward estimate minus the weighted lower-confidence cost
estimate. The orange arrow illustrates a cost reduction accompanied by lower
return. The arrows do not represent measured policy trajectories.}
\label{fig:method}
\end{figure}

Figure~\ref{fig:method} projects the exploratory action shift from
Section~\ref{sec:method} into estimated return-to-go and cost-to-go space. The
actual update is computed in action space. The figure is a conceptual
illustration and does not represent measured trajectories or guarantee that the
updated action improves both objectives.

\section{Additional Experimental Results}
\subsection{Component Ablations}
\label{app:ablations}

The mean-cost variant sets $\alpha_{\mathrm{pen}}=1$, causing the actor penalty
to average all $16$ predicted cost quantiles, and retains the exploratory action
shift. The no-shift variant retains the quantile-CVaR penalty and sets
$\delta=0$. All other method settings, including the task-specific multiplier,
remain unchanged.

All \emph{Hopper} policies are trained for $10^6$ interactions. Both ablations
remain tail-safe on every seed and obtain higher mean return than full OQ-SAC.
The full \emph{HalfCheetah} policy is trained for $5\times10^5$ interactions,
while its ablations use $10^6$. All three are tail-safe, but their returns are
not directly comparable because the training budgets differ.

\begin{table}[H]
\centering
\caption{Component ablations on \emph{HalfCheetah} and \emph{Hopper} with
three seeds and $d{=}25$. Values are means and standard deviations across
seeds. Every reported variant satisfies $\mathrm{CVaR}_{0.1}\le d$ on every
seed. Hopper uses $10^6$ training interactions for every row. Full OQ-SAC on
HalfCheetah uses $5\times10^5$, while its two ablations use $10^6$.}
\label{tab:ablation}
\setlength{\tabcolsep}{6pt}
\begin{tabular}{l cc cc}
\toprule
& \multicolumn{2}{c}{HalfCheetah} & \multicolumn{2}{c}{Hopper} \\
\cmidrule(lr){2-3}\cmidrule(lr){4-5}
Variant & $R\!\uparrow$ & $\mathrm{CVaR}\!\downarrow$ & $R\!\uparrow$ & $\mathrm{CVaR}\!\downarrow$ \\
\midrule
Full OQ-SAC & $2868_{\pm46}$ & $0_{\pm0}$ & $877_{\pm556}$ & $1.0_{\pm1.3}$ \\
Mean-cost variant & $2948_{\pm28}$ & $0_{\pm0}$ & $1246_{\pm398}$ & $0.5_{\pm0.4}$ \\
No-shift variant & $2908_{\pm32}$ & $0_{\pm0}$ & $1162_{\pm398}$ & $5.4_{\pm7.1}$ \\
\bottomrule
\end{tabular}
\end{table}

\subsection{Lower Expected-Cost Targets}
\label{app:sweep}

We test whether expected-cost baselines can produce a tail-safe policy when
trained with an internal cost target below the evaluation budget. Using training
seed zero, we evaluate SAC-Lagrangian and PPO-Lagrangian with
$d'\in\{25,10,5\}$. Here, $d'$ is the expected-cost target used during training.
Each final policy is evaluated for $100$ deterministic episodes against the
tail-safety budget $d=25$.

For each method and task, Table~\ref{tab:sweep} reports the highest-return
configuration satisfying $\mathrm{CVaR}_{0.1}\le25$. Lower targets produce
tail-safe SAC-Lagrangian policies on both tasks and a tail-safe PPO-Lagrangian
policy on \emph{Hopper}. No tested PPO-Lagrangian target is tail-safe on
\emph{HalfCheetah}. These single-seed results establish that lower internal
targets can produce safe operating points, but they do not establish reliability
across training seeds.

\begin{table}[H]
\centering
\caption{Highest-return tail-safe policy found in the single-seed expected-cost
target comparison. The internal training target $d'$ is shown in parentheses.
OQ-SAC is included as a three-seed reference and is not part of this comparison.
A dash indicates that no tested target produced a tail-safe policy.}
\label{tab:sweep}
\setlength{\tabcolsep}{4pt}
\begin{tabular}{l cc cc cc}
\toprule
& \multicolumn{2}{c}{OQ-SAC}
& \multicolumn{2}{c}{SAC-Lagrangian}
& \multicolumn{2}{c}{PPO-Lagrangian} \\
\cmidrule(lr){2-3}\cmidrule(lr){4-5}\cmidrule(lr){6-7}
Task & Return & CVaR & Return & CVaR & Return & CVaR \\
\midrule
HalfCheetah
& $2868$ & $0$
& $1693$ $(d'=10)$ & $0$
& -- & -- \\
Hopper
& $877$ & $1.0$
& $1000$ $(d'=5)$ & $0$
& $1319$ $(d'=5)$ & $0$ \\
\bottomrule
\end{tabular}
\end{table}

\subsection{Gaussian Distributional WCSAC Baseline}
\label{app:wcsac}

We implement WCSAC \citep{yang2021wcsac} with a Gaussian distributional cost
critic. The critic predicts the mean $\mu_C(s,a)$ and variance
$v_C(s,a)$ of the discounted cost return. Its upper-tail estimate is
\[
\rho_\alpha(s,a)
=
\mu_C(s,a)
+
\frac{\phi\!\left(\Phi^{-1}(1-\alpha)\right)}{\alpha}
\sqrt{v_C(s,a)},
\]
where $\phi$ and $\Phi$ are the standard normal density and distribution
functions. We use $\alpha=0.1$, and the actor is penalized by
$\lambda\rho_\alpha(s,a)$ with fixed multiplier $\lambda=5$. This state-action estimate is used during training.
Table~\ref{tab:wcsac} evaluates tail safety using empirical episode costs from
the shared $100$-episode evaluation. Results use three training seeds per
method, except OQ-SAC on \emph{Walker2d}, which uses five. WCSAC is tail-safe
on every seed for \emph{HalfCheetah} and \emph{Ant}. OQ-SAC is tail-safe on
every seed for \emph{HalfCheetah}, \emph{Ant}, and \emph{Hopper}, and on four
of five \emph{Walker2d} seeds.

\begin{table}[H]
\centering
\caption{OQ-SAC and WCSAC on four locomotion tasks with $d=25$. Values are
means and standard deviations across training seeds. Bold CVaR entries satisfy
the tail constraint on every seed. Underlined returns identify the
higher-return method among those satisfying this criterion.
$^{\dagger}$OQ-SAC is safe on four of five \emph{Walker2d} seeds. The remaining
seed has $\mathrm{CVaR}_{0.1}=28.3$.}
\label{tab:wcsac}
\setlength{\tabcolsep}{6pt}
\begin{tabular}{l cc cc}
\toprule
& \multicolumn{2}{c}{OQ-SAC} & \multicolumn{2}{c}{WCSAC} \\
\cmidrule(lr){2-3}\cmidrule(lr){4-5}
Task & $R\!\uparrow$ & $\mathrm{CVaR}\!\downarrow$ & $R\!\uparrow$ & $\mathrm{CVaR}\!\downarrow$ \\
\midrule
HalfCheetah & \underline{$2868_{\pm46}$} & \textbf{$0_{\pm0}$} & $2756_{\pm87}$ & \textbf{$0_{\pm0}$} \\
Ant         & \underline{$2656_{\pm329}$} & \textbf{$0.9_{\pm0.6}$} & $-683_{\pm230}$ & \textbf{$4.5_{\pm3.7}$} \\
Walker2d    & $1617_{\pm949}^{\dagger}$ & $10.7_{\pm10.1}$ & $481_{\pm64}$ & $34.3_{\pm21.8}$ \\
Hopper      & \underline{$877_{\pm556}$} & \textbf{$1.0_{\pm1.3}$} & $1016_{\pm391}$ & $179.8_{\pm165.7}$ \\
\bottomrule
\end{tabular}
\end{table}

\subsection{Per-Seed Return and Tail Cost}

\begin{figure}[H]
\centering
\includegraphics[width=0.70\textwidth]{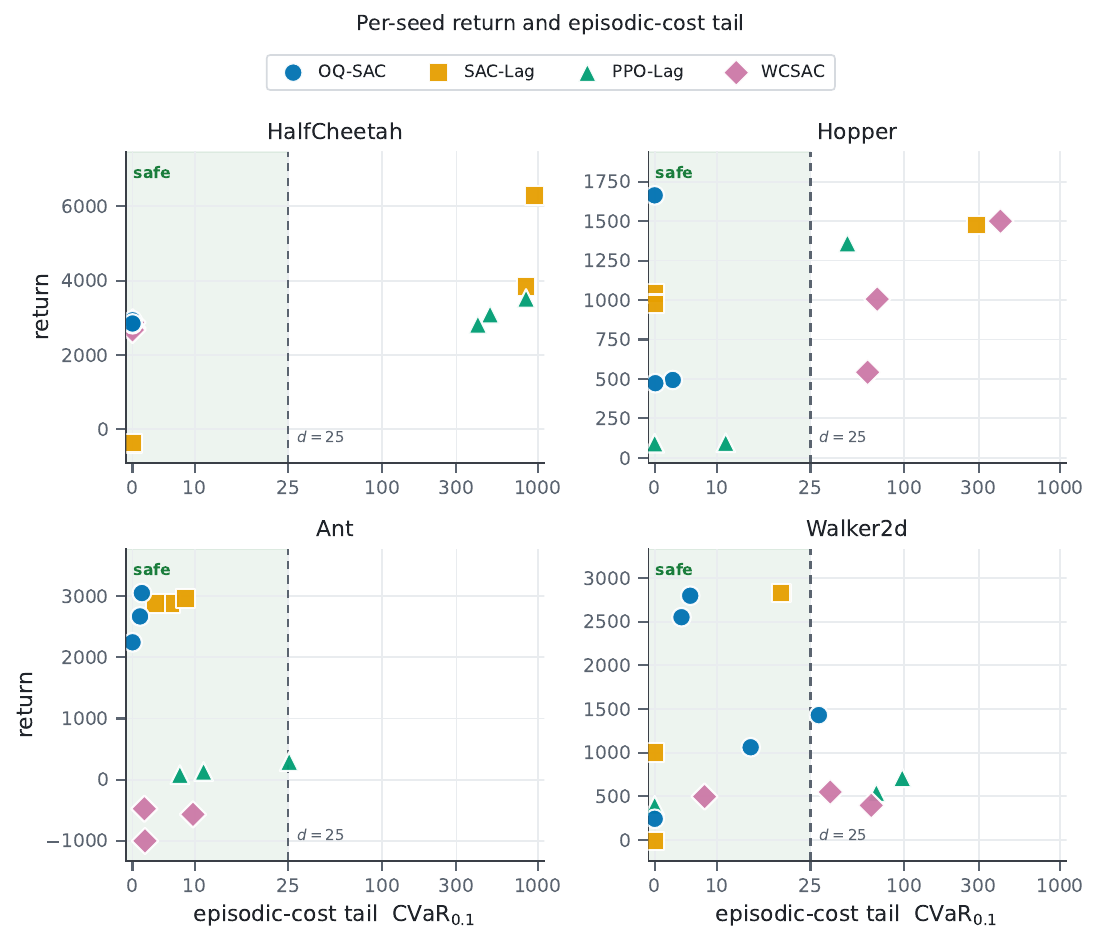}
\caption{Per-seed return and episodic-cost $\mathrm{CVaR}_{0.1}$ on the four
locomotion tasks. Each marker represents one trained policy evaluated for
$100$ deterministic episodes. The horizontal axis is linear within the budget
and logarithmic above it. The green region marks
$\mathrm{CVaR}_{0.1}\le d=25$.}
\label{fig:scatter}
\end{figure}

Figure~\ref{fig:scatter} shows the return and empirical
$\mathrm{CVaR}_{0.1}$ of every locomotion run reported in
Tables~\ref{tab:main} and~\ref{tab:wcsac}. Each marker represents one trained
policy evaluated for $100$ deterministic episodes. The figure exposes variation
across training seeds that is not visible from means and standard deviations
alone.

\section{Additional Evidence on Tail Controllability}
\subsection{Representative Episode-Cost Distributions}
\label{app:shape}

\begin{figure}[H]
\centering
\includegraphics[width=\textwidth]{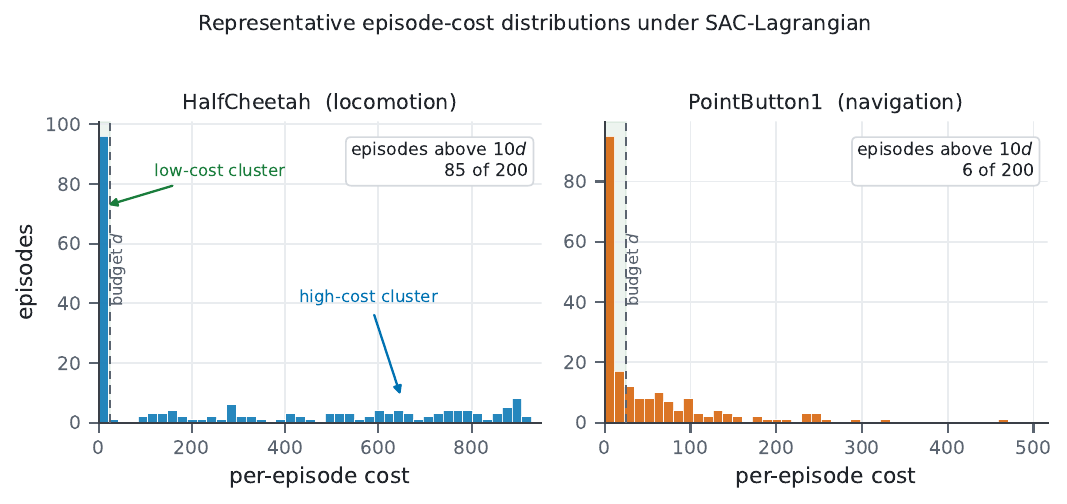}
\caption{Episode-cost distributions from $200$ deterministic evaluations of
representative SAC-Lagrangian policies. On \emph{HalfCheetah}, $85$ episodes
have cost above $10d$. On \emph{PointButton1}, $6$ episodes exceed this
threshold. The shaded region marks episode costs at or below $d=25$.}
\label{fig:separability}
\end{figure}

Figure~\ref{fig:separability} compares $200$ deterministic episodes from one
SAC-Lagrangian policy on \emph{HalfCheetah} and one on
\emph{PointButton1}. The HalfCheetah distribution contains separated low-cost
and high-cost clusters, with $85$ episodes exceeding $10d$.
The PointButton1 distribution is more diffuse, with $6$ episodes exceeding
$10d$. These examples illustrate different cost-distribution shapes but do not
establish that distribution shape determines tail controllability.

\subsection{Observed Return and Tail-Cost Operating Points}
\label{app:decouple}

\begin{figure}[H]
\centering
\includegraphics[width=\textwidth]{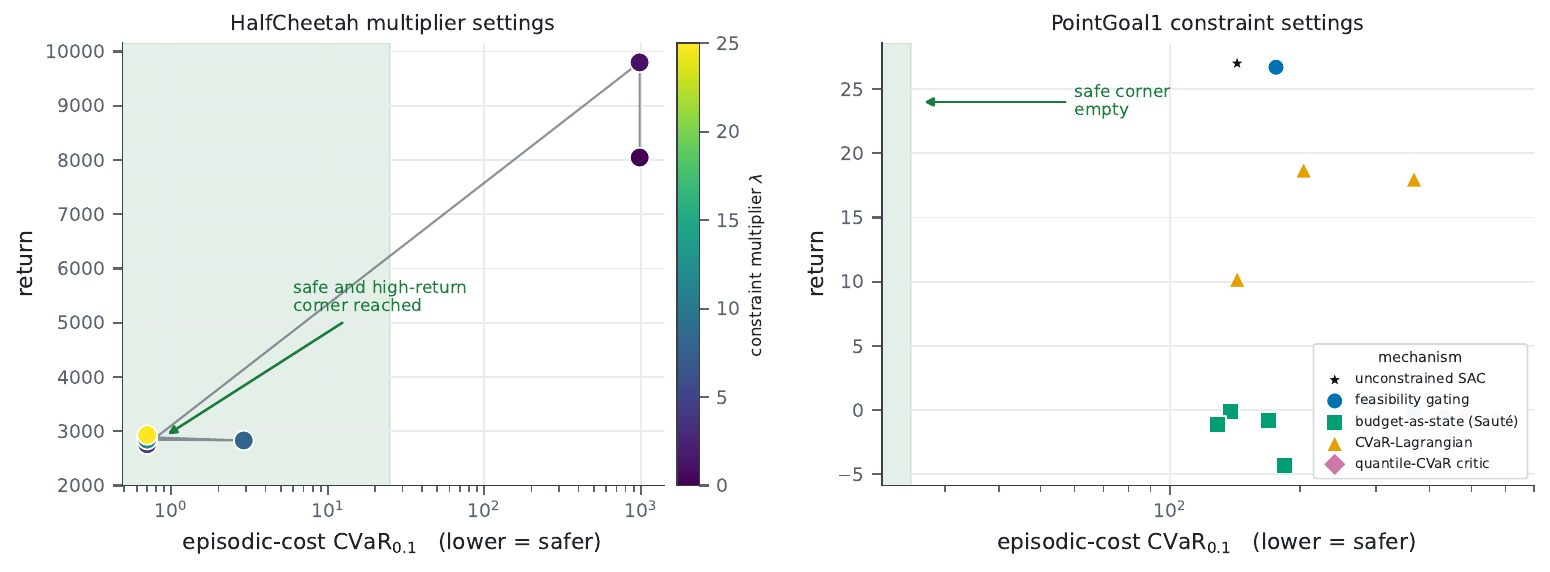}
\caption{Observed return and episodic-cost $\mathrm{CVaR}_{0.1}$ on two
representative tasks. The HalfCheetah panel shows seven OQ-SAC multiplier
settings, with the connecting line ordered by multiplier value. The PointGoal1
panel shows $17$ evaluated operating points across the tested constraint
families. The green region marks $\mathrm{CVaR}_{0.1}\le d=25$.}
\label{fig:decouple}
\end{figure}

Figure~\ref{fig:decouple} compares evaluated operating points on
\emph{HalfCheetah} and \emph{PointGoal1}. The HalfCheetah panel contains seven
OQ-SAC multiplier settings. Several settings satisfy
$\mathrm{CVaR}_{0.1}\le25$ with return near $2900$. The PointGoal1 panel
contains $17$ evaluated configurations from the four constraint families and
an unconstrained SAC policy. None reaches the tail-safe region. The complete
archive contains additional training traces that lack the standardized
post-training measurements required for this plot.

\subsection{Single-Seed Hazard-Count Comparison}
\label{app:density}

We vary the number of hazards in \emph{PointGoal1} while holding the OQ-SAC
configuration, training seed, fixed multiplier $\lambda=0.06$, and budget
$d=25$ unchanged. Each final policy is evaluated for $100$ deterministic
episodes. Table~\ref{tab:density} reports mean return, the fraction of episodes
within budget, mean return conditioned on an episode being within budget, and
$\mathrm{CVaR}_{0.1}$.

No tested hazard count is tail-safe. The one-, four-, and eight-hazard settings
retain mean return near $27$ while tail cost increases. Return falls to $6.8$
with sixteen hazards. The two-hazard result does not follow this trend. It has
$94\%$ safe episodes but $\mathrm{CVaR}_{0.1}=116.2$, showing that a high
frequency of safe episodes does not establish tail safety when the remaining
episodes incur severe costs.

\begin{table}[H]
\centering
\caption{Single-seed hazard-count comparison on \emph{PointGoal1}. Safe
episodes satisfy $C(\tau)\le d$. Safe-episode return is the mean return
conditioned on this event. Every policy is evaluated for $100$ deterministic
episodes, and none satisfies $\mathrm{CVaR}_{0.1}\le25$.}
\label{tab:density}
\begin{tabular}{lccccc}
\toprule
Hazards & Return & Safe\% & \makecell{Safe-episode\\return} & $\mathrm{CVaR}_{0.1}$ & CVaR/$d$ \\
\midrule
$1$  & $27.9$ & $90$ & $28.0$ & $40.9$  & $1.6$ \\
$2$  & $-0.4$ & $94$ & $-0.4$ & $116.2$ & $4.6$ \\
$4$  & $27.3$ & $62$ & $27.4$ & $82.3$  & $3.3$ \\
$8$  & $26.8$ & $34$ & $27.7$ & $176.9$ & $7.1$ \\
$16$ & $6.8$  & $63$ & $4.9$  & $397.2$ & $15.9$ \\
\bottomrule
\end{tabular}
\end{table}
\begin{figure}[H]
\centering
\includegraphics[width=\textwidth]{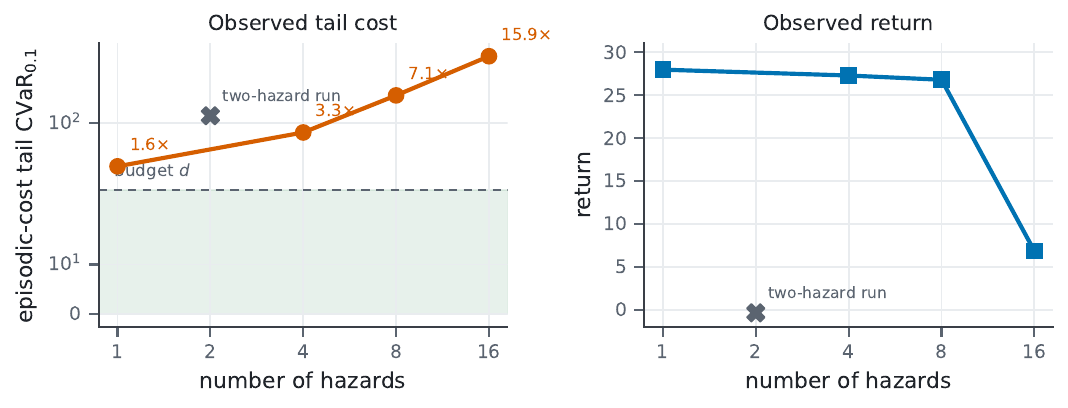}
\caption{Mean return and $\mathrm{CVaR}_{0.1}$ as the number of hazards varies.
The one-, four-, eight-, and sixteen-hazard results are connected in increasing
order. The two-hazard result is shown separately because it does not follow
this observed trend. The green region in the tail-cost panel marks
$\mathrm{CVaR}_{0.1}\le25$.}
\label{fig:density}
\end{figure}

\end{document}